\documentclass{article}

\usepackage{microtype}
\usepackage{graphicx}
\usepackage{subfigure}
\usepackage{booktabs} 

\usepackage[hyphens]{url}
\usepackage[hidelinks]{hyperref}
\hypersetup{breaklinks=true}
\usepackage{cite}

\usepackage{placeins}
\usepackage{makecell}

\usepackage[accepted]{icml2019}

\icmltitlerunning{Neural Networks for Fashion Image Classification and Visual Search}

\begin{document}

\twocolumn
[
\icmltitle{Neural Networks for Fashion Image Classification and Visual Search}



\icmlsetsymbol{equal}{*}

\begin{icmlauthorlist}

\icmlauthor{ \hspace{3pt} Fengzi Li }{} \textbar
\icmlauthor{  \hspace{3pt} Shashi Kant }{} \textbar
\icmlauthor{  \hspace{3pt} Shunichi Araki }{}  \textbar
\icmlauthor{  \hspace{3pt} Sumer Bangera }{} \textbar
\icmlauthor{  \hspace{3pt} Swapna Samir Shukla }{equal}

\end{icmlauthorlist}


\icmlkeywords{Machine Learning, ICML}

\vskip 0.3in
]



\printAffiliationsAndNotice{\icmlEqualContribution} %

\begin{abstract}
We discuss two potentially challenging problems faced by the ecommerce industry. One relates to the problem faced by sellers while uploading pictures of products on the platform for sale and the consequent manual tagging involved. It gives rise to misclassifications leading to its absence from search results.
The other problem concerns with the potential bottleneck in placing orders when a customer may not know the right keywords but has a visual impression of an image. An image based search algorithm can unleash the true potential of ecommerce by enabling customers to click a picture of an object and search for related products without the need for typing. In this paper, we explore machine learning algorithms which can help us solve both these problems.

\end{abstract}

\section{Introduction}
E-commerce has revolutionized the world of consumerism and unleashed a greater demand of products by providing a trouble-free buying experience and delivery to the user. We present two challenges confronting the industry - one from seller's perspective and the other from buyer's perspective. 

A seller who wishes to sell his product on an ecommerce platform has to upload pictures of his product and tag appropriate labels which render the product in search results. Human intervention makes this process prone to errors. Any misclassification thus caused may prevent the product from appearing in the search results thereby being responsible for less sales or no sale at all. Machine learning models can classify the images with high accuracy and prompt the sellers to do appropriate tagging.

There is another challenge that throttles the demand of a customer due to his lack of knowledge of the right keywords. In a typical e-commerce website, a user enters the keywords for the product he wants to purchase. Based on the keywords, the search algorithm matches it with product labels in its database and renders relevant products to the user. A user then explores among the search results to find the product which she wants to buy and places an order. A text-based search relies on the pre-requisite that the customer knows the product very well and knows the right keywords to punch into the search toolbar. However, this is not always the case. We come across many different things in our day-to-day life about which we are not aware. Such cases restrict our ability to search for the product on the e-commerce website. A visual search is an answer to this problem.

When customers undertake a visual search, they look for a product with a photo or other image instead of the keywords normally used in search engines. Shoppers can take a picture of something they want to buy, upload it to the visual search engine and immediately see visually similar items available to purchase. This idea is already being implemented by a lot of AI solution providers such as Visenze, Google Lens etc.

Any visual search algorithm is essentially a unsupervised problem wherein machine learning models can be deployed to learn features about the new images and search for similar products. Once the target image is uploaded to the visual search engine, more products of the similar features can be rendered as search results on the website. Typical visual search algorithms such as autoencoders can be used to generate the latent features of images. Besides, transfer learning using pre-trained deep neural network models are also employed to extract the embedding features of images. 

To summarise, this paper will pursue two broad objectives:
\begin{itemize}

\item \textbf{Image Classification}: To train different neural network models to learn from large sets of images of products from an e-commerce website. Use transfer learning with pre-trained models such as VGG19 to do image classification. 

\item \textbf{Image Search}: Use autoencoders and cosine similarity to identify similar images. 
\end{itemize}

\section{Data and Exploratory Data Analysis}

We use real life fashion images from an Indian ecommerce website. The Fashion Product Images (Small)  dataset\footnote{\url{https://www.kaggle.com/paramaggarwal/fashion-product-images-small}} is readily available on Kaggle and therefore, does away the need for scraping images from website. Another advantage of using this data is that it has color images which makes it closer to a real-life business problem as opposed to the Fashion MNIST data from Kaggle that has greyscale images only. The dataset can be directly downloaded into the Google Collaboratory (GC) environment using Kaggle API. GC offers seamless importability of Kaggle data and provides free GPU/ TPU resources to employ computing-intensive convolutional neural network (CNN) algorithms for image classification. It also includes pre-installed python libraries like TensorFlow which make CNN modeling of images convenient. 
In addition to professionally shot high resolution product images, the dataset also has multiple label attributes describing the product which was manually entered while cataloging. We choose the small image dataset which contains the same set of images but with lower resolution to save on computational resources and run-time. 

\begin{figure}[t]
\vskip 0.2in
\begin{center}
\centerline{\includegraphics[width=\columnwidth]{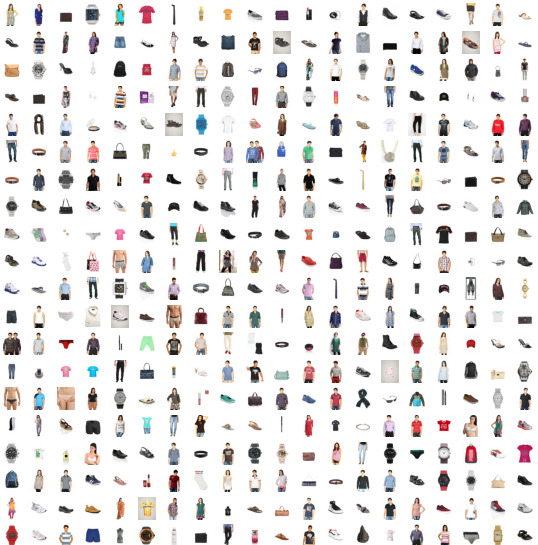}}
\caption{Random selections from Fashion Product dataset on Kaggle.}
\label{icml-historical}
\end{center}
\vskip -0.2in
\end{figure}

The Fashion Product Metadata \texttt{styles.csv} has 44,424 images along with product description and attributes.  The images folder consists of 44,441 images. The images are provided in jpg format with each image of size 80 x 60 pixels in 3 color channels.  Each product image as identified by its numeric id can be mapped to its metadata in \texttt{styles.csv}. Upon merger of the two, we find 44,419 instances where image IDs are matched with the IDs in metadata file. We consider these set of images for modelling. 

The final dataset has fully labeled attributes such as Gender, Master Category, Sub Category, Article Type, Season etc. We use three schemes for product classification namely - a concatenated Gender and Master category, Sub Category and Article Type; as they are cleanly labelled without any missing values.

\section{Data Pre-processing}

\begin{figure*}[t]
\vskip 0.2in
\begin{center}
\centerline{\includegraphics[width=\linewidth]{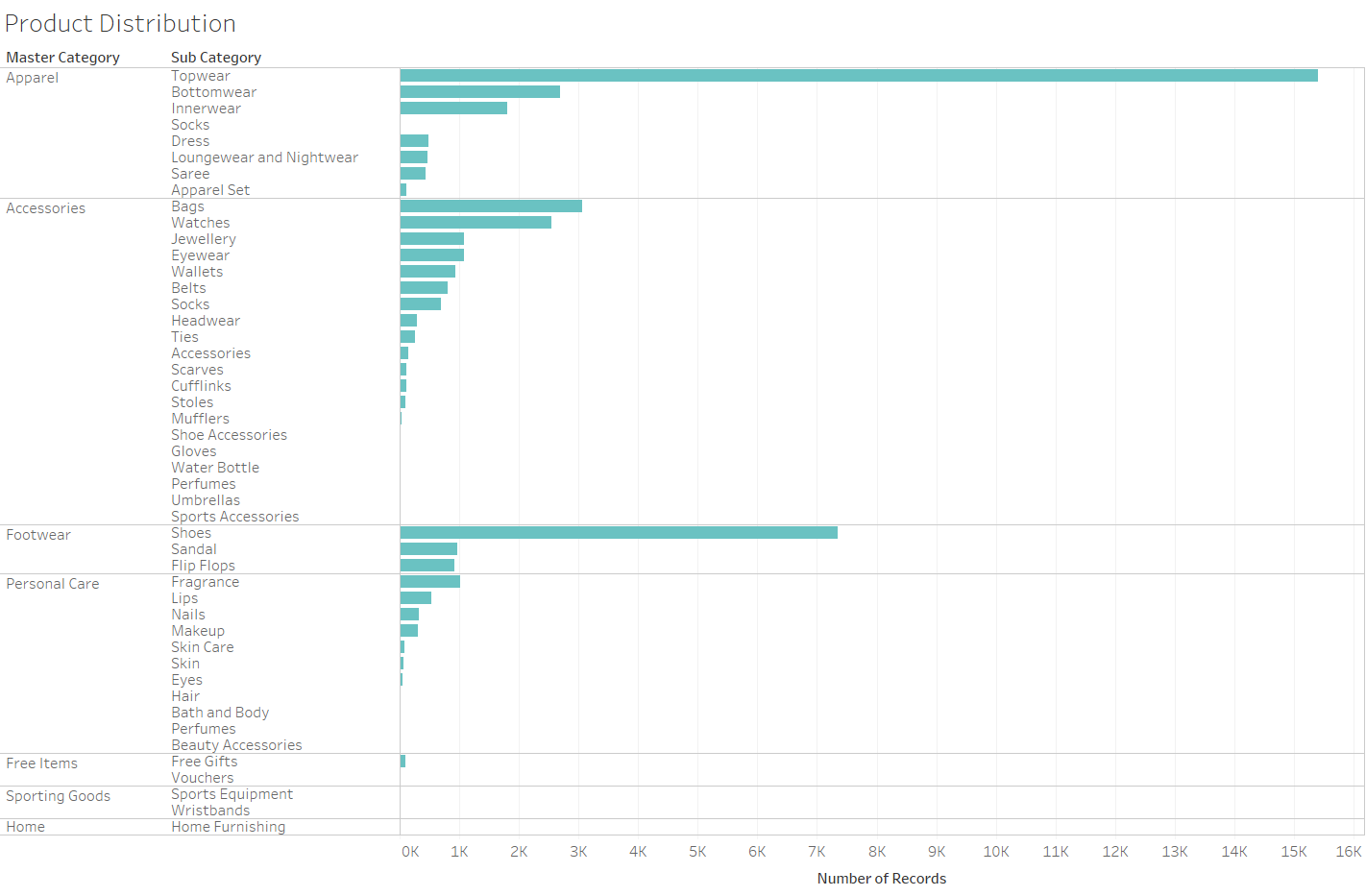}}
\caption{Product distribution before filtering.}
\label{Product_Distribution}
\end{center}
\vskip -0.2in
\end{figure*}

The Fashion Product dataset is a clean labelled dataset, however, it is highly unbalanced for the purpose of using machine learning algorithms to do image classification (Figure ~\ref{Product_Distribution}). We follow a two-tier data pre-processing to address the data imbalance.

\subsection{Data Filtration}

Out of the multiple image attributes, we choose three image attributes for our classification objective, namely – concatenated Gender \& Master Category; Sub-Category; and Article Type. There are a lot of classes which have very few images. For example, articleType of Wristbands only has 7 products while Casual Shoes has 2845 different products.  Machine learning algorithms tend to get biased by the classes having majority which leads to identifying even the minority classes as majority. In such situations, the misclassification of minority classes does not lead to a sharp drop in accuracy and therefore may give a false perception of the model performing well. In order to avoid such misclassification, balancing the datasets is necessary. To achieve this, we remove the classes which have less than 500 images. A snapshot of the filtered dataset is presented in Table \ref{sample-table}.

\begin{table}[b]
\caption{Balancing of data by filtering classes that have less than 500 images.}
\label{sample-table}
\vskip 0.15in
\begin{center}
\begin{small}
\begin{sc}
\resizebox{\columnwidth}{!}{%
\begin{tabular}{lcccr}
\toprule
    Data Set & \multicolumn{2}{p{8.11em}}{\thead{Before \\ Filtering}} & \multicolumn{2}{p{8.11em}}{\thead{After \\ Filtering}} \\
    \midrule
    \multicolumn{1}{r}{} & \multicolumn{1}{p{4.055em}}{classes} & \multicolumn{1}{p{4.055em}}{images} & \multicolumn{1}{p{4.055em}}{classes} & \multicolumn{1}{p{4.055em}}{images} \\
    Gender \& Master Category &  45     &  44,419     & 15      & 44,018 \\
    Sub Category & 45      &44,419       & 12      & 40,835 \\
    Article Type & 142      & 44,419      & 23      & 34,697 \\
    \bottomrule
\end{tabular}
}
\end{sc}
\end{small}
\end{center}
\vskip -0.1in
\end{table}

Most of the excluded classes contain very few samples and therefore may not have a significant business value. Moreover, it is deemed that incorporating classes with extremely few samples will not give good prediction even with imbalanced class handling techniques such as data augmentation and SMOTE. In other words, there is a trade-off between including as many classes as possible and maintaining quality of the input data in order build a limited yet a good model.

\subsection{Data Augmentation}
After the selection of classes, we do further rebalancing by using data augmentation technique with the help of \texttt{ImageDataGenerator} class. The Keras deep learning library provides the ability to use data augmentation automatically when training a model \cite{brownlee_how_2019}. 
We use rotation, horizontal flipping, shifting and zoom parameters to augment the dataset. This function only returns the randomly transformed data during training time.

\subsection{Weight Balancing}

We also experimented with the weighted class approach. Weight balancing is another approach to address the issue of imbalance in training datasets. It balances the data by assigning higher weights to the minority classes \cite{bhattacharya_how_2020}. Misclassification of minority classes has a higher penalty in the loss function, thereby, ensuring that the model doesn’t identify it as a majority class. However, we observed that weight balancing does not improve accuracy by a big margin. So we do not include it in the final model for the sake of simplicity.

\subsection{SMOTE}

SMOTE is an oversampling technique that generates synthetic samples from the minority class \cite{bhattacharya_how_2020}. It is commonly used with structured 2D data but is not so popular when dealing with images as it requires images to be reshaped to 2D array. Since images have three channels, reshaping results in a large array which creates issues owing to computational limits. SMOTE was only applied for small subset of data for testing purpose while it is not used in the proposed models.

\subsection{Under-sampling}
It does not make full use of available dataset. Since we only have limited number of images, we do not use this in our proposed model.

\section{Models \& Methodology}

Precise labeling of products is pivotal to businesses that rely on an e-commerce model as customer experience becomes a key driver to generate revenues from sales. Conversely, incorrect labeling may lead to undesirable search results, adversely affecting the customer experience. To circumvent such scenarios, it is imperative for the business to ensure that the products are appropriately categorized across classes such as gender, product category, product type, etc. Besides, an efficient and accurate image classification model can help e-commerce to automatize online product management, reducing operation cost and discrepancies in the system.
Our model focuses on alleviating these problems that e-commerce businesses, in all likelihood, experience as their business and volume of products grow over time. To achieve this, we train our model at three different levels of hierarchy in the product classification, viz. product categorization based on Gender and Master category combined (least granular), Sub-category and Article-type (most granular). As discussed earlier, we exclude categories that have less than 500 samples at each stage to ensure the model uses a sufficient number of training samples. Moreover, since training deep convolutional neural network models is computationally expensive when the dataset is large, we apply transfer learning using advanced pre-trained models such as VGG-19 and Inception V3. 
Another important feature being explored and invested in by the e-commerce businesses is the visual search feature which provides an exceptional and innovative way to search for products faster and conveniently. To implement this feature, the e-commerce business should have model to efficiently identify and search the product requested by the customer. However, a simple visual search feature will be insufficient if the requested product is unavailable in the inventory. Therefore, our model aims to visual search feature with the capability to recommend products similar to the customer’s request. For this, we have applied two approaches – first, building an autoencoder based on convolutional neural network (CNN) architecture, and second, using the pre-trained ResNet-50 model.

In a CNN model, the low-level features such as lines are captured by the convolutional layers closer to the input layer whereas the middle layers learn the abstract and more complex features combined with other lower-level features. The layers closer to the output layer are significantly important as they interpret the extracted features for the classification task.
Since our dataset is different from the \texttt{ImageNet} dataset, we only use VGG-19 and Inception V3 for integrated feature extraction. In other words, we integrate these pre-trained models into our network by freezing their layers during training. 
Furthermore, to optimize the model development process, we first focussed our interest in one master category i.e. Footwear, consisting of 9,219 samples across 8 classes (article types). Based on the results of this pilot testing, accuracy 76.1\%, we decided to scale up the models for the entire dataset. 

We firstly use the random sampling to get training and testing dataset and secondly use the \texttt{train\_test\_split} in ImageDataGenerator to obtain the training and validation dataset for model training.
For each granular level modeling, 80\% of the subset of data is used during the model training process while the remaining 20\% of data is reserved for testing the model. Additionally, within the training dataset, we reserve 20\% samples as the validation set.

\subsection{Transfer Learning}

Deep convolutional neural network models require significant training time when the dataset is exceptionally large. In this study, we encounter the same issue while handling our dataset even though it consists of images of lower resolution.  Therefore, it is necessary to employ alternative model building strategies such as transfer learning, wherein we extract and reuse the model weights from previously trained models developed using a mega high-resolution image dataset.  This method has two significantly important advantages. Firstly, the time required to train the neural network is much lower and secondly, the model thus developed is more generalizable. In our neural network models, we have integrated the two pre-trained models for the weight initialization step since the original datasets, on which these models were developed, consist of a higher number of labels than our dataset. The original dataset refers to the ImageNet Large Scale Visual Recognition Challenge which contains over 14 million images of everyday objects across 1,000 classes. In addition to this, the pre-trained models are easily accessible through the Keras library in Python

\subsection{Image classification using transfer learning}

To fully understand the model building process, it is important to comprehend the underlying architecture of the pre-trained models, viz. VGG-19, Inception V3 and ResNet-50.

\subsubsection{VGG-19}

The architecture of VGG-19 consists of 19 deep layers and an input layer of size $224 \times 224$ \cite{simonyan_very_2015}. The only preprocessing performed by us was the subtraction of the mean RGB value from each pixel, computed over the whole training set. Additionally, we also used kernels of size $3 \times 3$ and convolution stride fixed to 1 pixel to capture the entire notion of the input image. The spatial padding was designed to preserve the spatial resolution of the input image. The network uses $3 \times 3$ stacked convolutional layers in increasing depth, and max-pooling over a $2 \times 2$ pixel window with stride 2 to reduce the volume size. Following this, there are three fully connected layers, two containing 4096 nodes and the third with 1000 channels for 1000 way \textit{ILSVRC} classification and subsequently the final Softmax function layer. We train all the hidden layers with Rectified linear unit (ReLu) to introduce non-linearity.

To build our first model, we use this VGG-19 architecture with modifications to overcome the problems that may arise because the dataset is different from the ImageNet dataset. For this, we re-train the last 5 layers of VGG-19 and lock the remaining layers with the pre-trained weights. To achieve the target number of classes, we add three dense layers and the last layer using Softmax function to derive the class probability outputs. Detailed architecture can be found in the Appendix.

In addition to the above, we use regularization to facilitate the construction of a deep and accurate neural network model while maintaining good generalizability. Some regularization techniques include parameter norm penalties, early stopping, weight sharing, drop-out, data augmentation, noise injection and batch normalization. In our present study, we apply early stopping, reduce learning rate on plateau, drop-out at the newly added penultimate dense layer and data augmentation. The weight sharing stage is implicitly incorporated through convolution in the pre-trained architecture.

\subsubsection{Inception V3}

Inception V3, the third iteration of the inception architecture, uses 48 deep layers and stacks 11 inception modules, each consisting of pooling layers and convolutional filters \cite{ding_deep_2019}. The default input size for this model is 299x299. Moreover, it consists of three fully connected layers of varied sizes, \textit{viz.} 1024, 512 and 3 which are added to the ultimate concatenation layer. Like VGG-19, it uses Rectified linear unit (ReLu) as the activation function.

\subsection{Visual search using autoencoders and transfer learning}

To achieve the objective of building a successful visual search model with the ability to recommend similar products, we use two powerful models – CNN-based Autoencoders and ResNet-50.

\subsubsection{CNN based autoencoder}

Autoencoders can be used to find similar images and thus it is useful for image search as well as product recommendation. Autoencoders are neural networks which comprises of both an encoder and a decoder. Encoder is used to generate embedding features which is dimension reduced and dense to represent the most pertaining and essential features of input, i.e. images for the current study. Once an autoencoder model is trained,  encoder function can be used to predict the embedding features for each image which will be reshaped in a vector form. Autoencoder model training is a self-supervised learning with the input and output being the same image itself. Once the embedding features are obtained from encoder, cosine similarity scores can be calculated and the most similar images can be retrieved according to the score ranking. 

We build a 14-layer CNN architecture as the autoencoder. The first 7 layers is the encoder. Embedding features are extracted a the 7th layer with output size of (8,8,8). After flattening, the embedding feature vector size becomes 512 which is much dense compared to the original size of 14,400 for original image size of (80,60,3).

\subsubsection{ResNet based autoencoder}

We implement an alternative model building method using transfer learning because the aforementioned CNN model is slow to train. To get embeddings to identify the features of fashion items, we apply the model without any adjustments, which have one layer to convert image data into flattening vector. We use ResNet-50’s embeddings to identify similar items and as the input of the model, containing a reduced dimensionality but with much semantic information for the items.

Residual Network, or ResNet, uses a network-in-network architecture also known as micro-architecture modules. It uses the principle of “identity shortcut connection” to overcome the “vanishing gradient” problem which make the neural network hard to train. ResNet-50 consists of 50 deep layers, a default input size of $224 \times 224$ and more than 23 million parameters to train. It comprises of a 5-staged model where each stage consists of a convolution block and an identity block, each containing 3 convolution layers. Moreover, ResNet-50 uses global average pooling unlike fully connected layers used in VGG-19 and Inception V3, which further reduces the model size in comparison.

\subsection{Selection of accuracy metric}

To evaluate our classification models, we use confusion matrix and precision-recall curve. Confusion matrix plots summarize how each product class is categorized. We present a normalized confusion matrix plot where values in the diagonal axis indicate the percentage of products that are correctly classified while other values outside the diagonal axis indicate the percentage of products that are misclassified.

For the results of visual search, we were only able to do able qualitative visual inspection to assess the accuracy. This may be explored in future.

\section{Results}

\subsection{Image Classification}
\subsubsection{VGG19}
The results of the VGG-19 based product categorization model have been summarized in the table below (Table \ref{Results}). The model was tested for each granularity level in the product hierarchy. We observe good performance overall while the model accuracy is the best when the model is trained to predict the sub-categories of the products. It is worth noting the model performance in predicting the article type even though many of the articles are very similar to each other. 

\begin{table}[ht]
\caption{VGG 19 model performance.}
\label{Results}
\vskip 0.15in
\begin{center}
\begin{small}
\begin{sc}
\resizebox{\columnwidth}{!}{%
\begin{tabular}{p{7.665em}ccp{7em}}
    \toprule
    {\thead{Product \\ Category}} & {\thead{Val \\Accuracy}} & {\thead{Test \\ accuracy}} & {Epochs, Runtime} \\
    \midrule
    Gender  \& Master Category & 0.871 & 0.884 & 50 Eps, 3054s \\
    Sub-Category & 0.957 & 0.966 & 33 Eps, 1907s \\
    Article Type & 0.841 & 0.873 & 18 Eps, 1173s \\
    \bottomrule
    \end{tabular}%
}
\end{sc}
\end{small}
\end{center}
\vskip -0.1in
\end{table}

\begin{figure}[hb]
\vskip 0.2in
\begin{center}
\centerline{\includegraphics[width=\columnwidth]{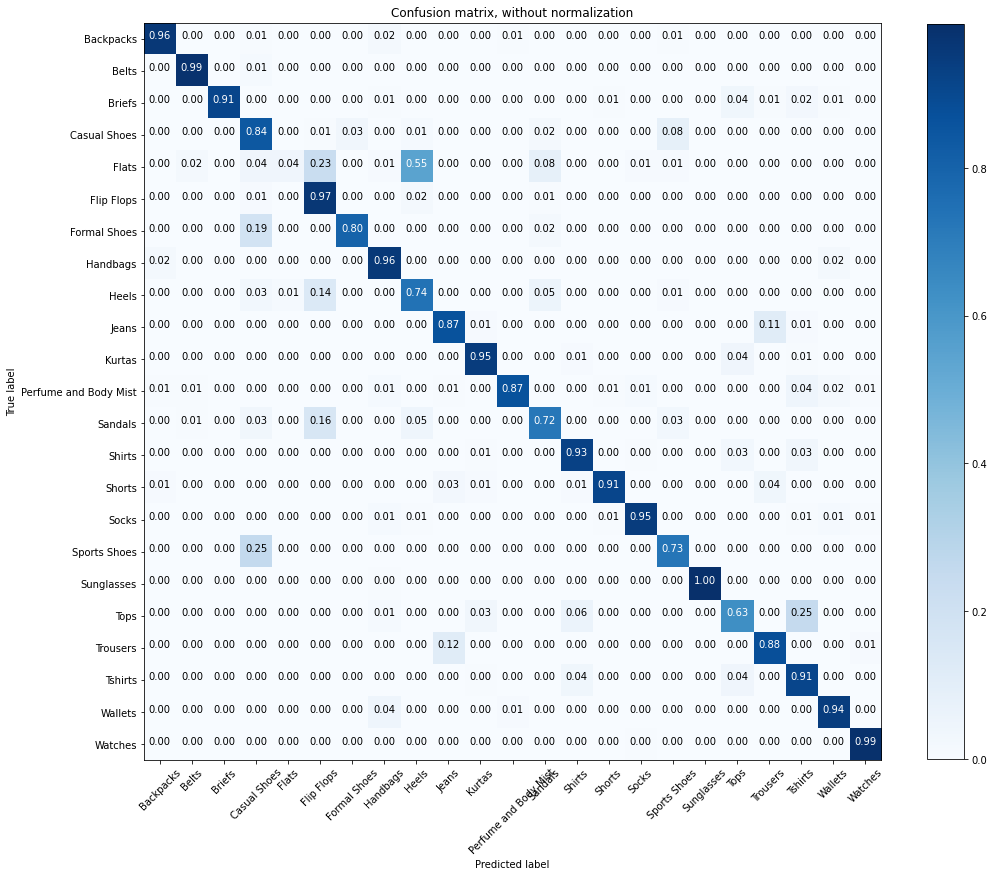}}
\caption{Confusion matrix from classification by article type.}
\label{confusion_matrix}
\end{center}
\vskip -0.2in
\end{figure}

To understand the model performance further in detail, we use confusion matrix and precision-recall curve to evaluate the classification predictions. For illustration, only the results from the article type classification model are shown in Figure \ref{confusion_matrix}.

\begin{figure}[hb]
\vskip 0.2in
\begin{center}
\centerline{\includegraphics[width=\columnwidth]{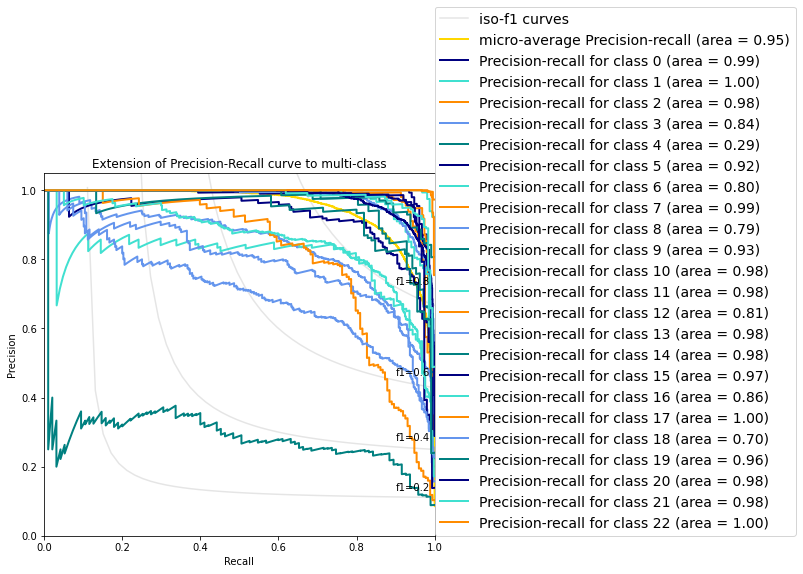}}
\caption{Precision Recall Curve from classification by article type.}
\label{precision-recall-curve}
\end{center}
\vskip -0.2in
\end{figure}

As observed, the accuracy of product classification is good overall, but some expected misclassifications are apparent from the confusion matrix. For example, 55\% of Flats are misclassified as Heels and 25\% Sports Shoes are misclassified as Casual Shoes, 25\% of Tops are misclassified as T-Shirts. These misclassifications are within a reasonable range as these classes are very similar to each other. In addition, Flats category has the least number of samples as compared to other classes obtained from the pre-processing stage. The lack of sufficient samples for some product classes may be a possible reason for poor classification as the model is inadequately trained to learn all the features. Another possible reason for misclassifications may arise due to incorrect labels in the training data which may lead to spurious results even if a model used is very sophisticated. Therefore, it is imperative for the e-commerce firm to ensure high-quality training data to maximize the advantages of neural networks.
The precision-recall-curve for article type is shown in Figure \ref{precision-recall-curve}. It demonstrates more explicitly the predictions made by the model for each class. Precision-recall for class 4 with an area of 0.29 belong to Flats class and this is consistent with the confusion matrix plot.

\subsubsection{Inception V3}
The Inception V3 model was trained on Article Type sub-dataset.All the layers in the model were used except for the last fully connected layer as it is specific to the ImageNet competition. Our optimizer is RMSprop with a learning rate of 0.0001. Since we’re trying to predict classes, we use categorical crossentropy as our loss function. The model was retrained on all the layers, in addition to an average pooling layer, two dense layers and, a drop out layer to account for overfitting.

\begin{figure}[t]
\vskip 0.2in
\begin{center}
\centerline{\includegraphics[width=\columnwidth]{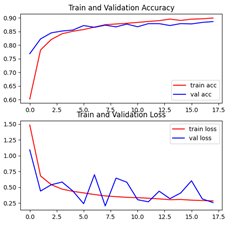}}
\caption{Accuracy performance of Inception V3 model.}
\label{inception}
\end{center}
\vskip -0.2in
\end{figure}

While the validation accuracy peaked at 0.88654, the Test accuracy dropped down to 0.18 (Figure \ref{inception}). This is not due to overfitting, as making all 48 layers of Inception V3 untrainable yielded even poorer results. One possible reason could be due to resizing of Input images from 80 *60 to 75*75. 
Inception V3 requires the minimum size of images to be 75*75. Converting an 80*60 image to an even higher resolution like 139*139 or 295*295 renders the image unsuitable for training, and yields even poorer validation accuracy. It would seem that presence of 48 layers makes customisation difficult and the use of batch normalization layers, further makes the process of training new layers, freezing convolution layers and subsequent fine-tuning of convolution layers after training of new layers difficult.

\subsection{Visual Search}

\begin{figure*}[ht]
\vskip 0.2in
\begin{center}
\centerline{\includegraphics[height = 8 cm, width=\linewidth]{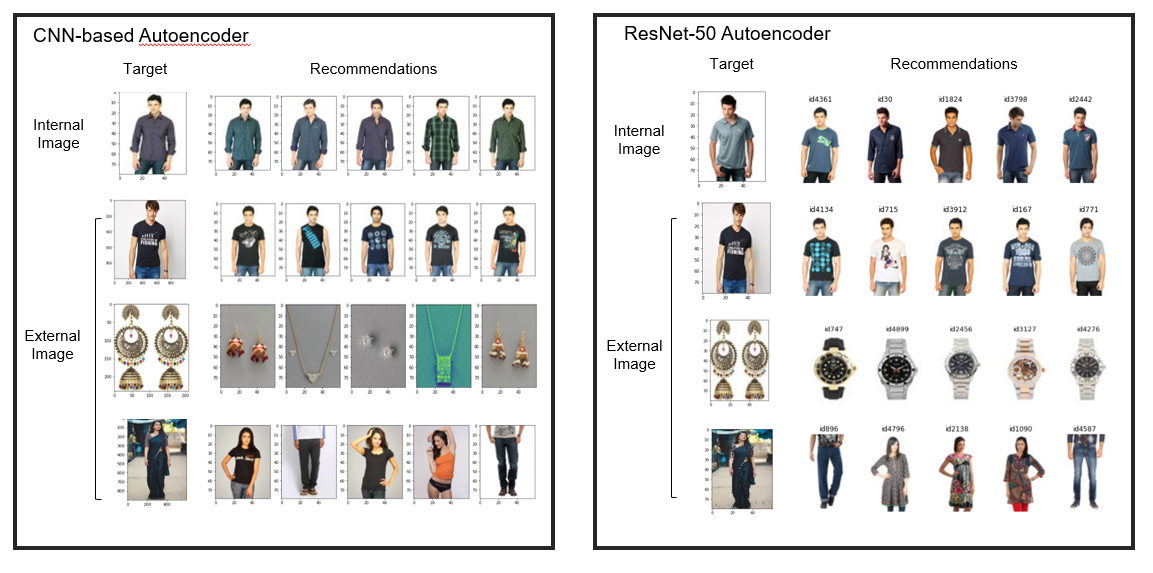}}
\caption{Visual search results from CNN autoencoders and ResNet-50 autoencoders.}
\label{purple-shirt}
\end{center}
\vskip -0.2in
\end{figure*}

To test our visual search models, we use the cosine similarity approach. Cosine similarity measures the similarity between two non-zero vector arrays in a vector space by calculating the cosine of the angle between these two vectors. The outcome is neatly bounded in [0,1] with 1 indicating two vectors having the same orientation as ‘similar’ while 0 indicating orthogonal as ‘dissimilar. 
Furthermore, the visual search models are evaluated in a qualitative manner, namely: 1) random selection of a target image to get recommendations from the current dataset; 2) comparing the recommended image labels (gender, masterCategory, subcategory and articleType) against the target image; 3) using external images as target and retrieve the recommended images from the current dataset. 
In the CNN autoencoder and ResNet-50 embedding models, the top 5 matching images are recommended during the visual search process. Figure \ref{purple-shirt} shows the results of image search using the internal and external images. The leftmost image in each box is the target image while the five images on its right are the recommendation results using the autoencoder model. 

In order to evaluate the generalization of our models, we use external images obtained from the web as the targets to retrieve the recommended images from the current dataset. The results of the visual search using the two types of autoencoders are placed in Appendix. We have only included relevant search results for tshirts, earrings and saree so as to highlight the model accuracy and shortcomings both to provide a holistic view. Therefore, these figures are only a representative set of our testing results that help acknowledge important model capabilities. As stated earlier, devising the performance metric for visual search results can be a project for future. 

\section{Limitations}
The accuracy in the current study might be limited due to the use of low resolution images or the mislabelling of products. On one hand, low resolution images are faster to train the model but render a compromise in terms of accuracy. On the other hand, the usage of high resolution images in the model may cause longer runtime in training the models albeit with higher accuracy. The tradeoff between runtime and accuracy is something which the business has to decide for itself. 

Machine learning models can be trained to achieve high accuracy levels but we need to be mindful of the imbalance in data. It is desirable to have sufficient number of images in each class. As much as this may sound a reasonable requirement, it is hardly the case in practice. A lot of products have only limited number of images which pose a challenge in training the models in an unbiased way. This is often the case when a brand new product is going to be launched in the market. ML models cannot be trained to identify such instances.

\section{Conclusion and Way Forward}

Machine learning models find a very interesting use case in enabling visual search which solves the problem of searching products when the right keywords are not known to the user. It enhances customer experience by enabling user  to search for similar products just by clicking a photo of the product at hand. 

Machine learning models can be also be very useful in enhancing seller experience in listing their products on the platform.  Sellers can upload photos of their products and automated image-to-text machine learning algorithms can generate appropriate tags to label them. This can reduce the inaccuracies in labelling products which often affect the demand adversely as the products are not rendered correctly in the search results. To achieve this, CNN models need to be combined with NLP techniques such as Word2vec to predict text information from the image data and its features. 

Another possible use case for image classification could be the identification of counterfeit products \cite{noauthor_how_nodate}. An extensive study of the attributes of a brand’s logo such as design, colors, position, etc. can help in the identification of fake products.

 With the rapid advancements in computational power and machine learning algorithms, we may even employ a generative adversarial network (GAN) to come up with new designs of fashion accessories and curtail our dependence on human creativity. Although it is said that training a GAN model is very hard, GAN models in the fashion industry can have significant business value in the near future.

\nocite{noauthor_importance_2017, seif_handling_2019, synced_visual_2018,
muses_shopping_2019, noauthor_what_nodate}

\bibliography{example_paper}

\begin{thebibliography}{10}
\providecommand{\natexlab}[1]{#1}
\providecommand{\url}[1]{\texttt{#1}}
\expandafter\ifx\csname urlstyle\endcsname\relax
  \providecommand{\doi}[1]{doi: #1}\else
  \providecommand{\doi}{doi: \begingroup \urlstyle{rm}\Url}\fi

\bibitem[noa()]{noauthor_what_nodate}
What is visual search: how retailers can use it to enhance the customer
  experience.
\newblock URL \url{https://www.shopify.com.sg/retail/what-is-visual-search}.

\bibitem[noa(2017)]{noauthor_importance_2017}
The importance of visual search in your ecommerce website, July 2017.
\newblock URL \url{https://curatti.com/visual-search-ecommerce/}.

\bibitem[Bhattacharya(2020)]{bhattacharya_how_2020}
Bhattacharya, A.
\newblock How to use {SMOTE} for dealing with imbalanced image dataset for
  solving classification problems, February 2020.
\newblock URL
  \url{https://medium.com/swlh/how-to-use-smote-for-dealing-with-imbalanced-image-dataset-for-solving-classification-problems-3aba7d2b9cad}.

\bibitem[Brownlee(2019)]{brownlee_how_2019}
Brownlee, J.
\newblock How to configure image data augmentation in keras, April 2019.
\newblock URL
  \url{https://machinelearningmastery.com/how-to-configure-image-data-augmentation-when-training-deep-learning-neural-networks/}.

\bibitem[Ding et~al.(2019)Ding, Sohn, Kawczynski, Trivedi, Harnish, Jenkins,
  Lituiev, Copeland, Aboian, Mari~Aparici, Behr, Flavell, Huang, Zalocusky,
  Nardo, Seo, Hawkins, Hernandez~Pampaloni, Hadley, and Franc]{ding_deep_2019}
Ding, Y., Sohn, J.~H., Kawczynski, M.~G., Trivedi, H., Harnish, R., Jenkins,
  N.~W., Lituiev, D., Copeland, T.~P., Aboian, M.~S., Mari~Aparici, C., Behr,
  S.~C., Flavell, R.~R., Huang, S.-Y., Zalocusky, K.~A., Nardo, L., Seo, Y.,
  Hawkins, R.~A., Hernandez~Pampaloni, M., Hadley, D., and Franc, B.~L.
\newblock A deep learning model to predict a diagnosis of alzheimer disease by
  using 18f-fdg pet of the brain.
\newblock \emph{Radiology}, 290\penalty0 (2):\penalty0 456--464, 2019.
\newblock ISSN 1527-1315.
\newblock \doi{10.1148/radiol.2018180958}.

\bibitem[Muses(2019)]{muses_shopping_2019}
Muses, M.
\newblock Shopping with your camera: how visual search is transforming
  ecommerce, August 2019.
\newblock URL
  \url{https://towardsdatascience.com/shopping-with-your-camera-how-visual-search-is-transforming-ecommerce-1bee5877994e}.

\bibitem[Nvidia(2018)]{noauthor_how_nodate}
Nvidia.
\newblock How ai is helping consumer brands detect and eliminate counterfeit
  products – nvidia developer news center, 2018.
\newblock URL
  \url{https://news.developer.nvidia.com/how-ai-is-helping-consumer-brands-detect-and-eliminate-counterfeit-products/}.

\bibitem[Seif(2019)]{seif_handling_2019}
Seif, G.
\newblock Handling imbalanced datasets in deep learning, May 2019.
\newblock URL
  \url{https://towardsdatascience.com/handling-imbalanced-datasets-in-deep-learning-f48407a0e758}.

\bibitem[Simonyan \& Zisserman(2015)Simonyan and Zisserman]{simonyan_very_2015}
Simonyan, K. and Zisserman, A.
\newblock Very deep convolutional networks for large-scale image recognition.
\newblock \emph{arXiv:1409.1556 [cs]}, April 2015.
\newblock URL \url{http://arxiv.org/abs/1409.1556}.
\newblock arXiv: 1409.1556.

\bibitem[{Synced}(2018)]{synced_visual_2018}
{Synced}.
\newblock Visual search is revolutionizing e-commerce, September 2018.
\newblock URL
  \url{https://medium.com/syncedreview/visual-search-is-revolutionizing-e-commerce-b27a37dbd296}.

\end{thebibliography}
\bibliographystyle{icml2019}







\end{document}